\documentclass{article}
\usepackage[preprint,nonatbib]{nips_2018}
\usepackage[utf8]{inputenc} % allow utf-8 input
\usepackage[T1]{fontenc}    % use 8-bit T1 fonts
\usepackage{hyperref}       % hyperlinks
\usepackage{url}            % simple URL typesetting
\usepackage{booktabs}       % professional-quality tables
\usepackage{amsfonts}       % blackboard math symbols
\usepackage{nicefrac}       % compact symbols for 1/2, etc.
\usepackage{microtype}      % microtypography
\usepackage{xcolor}
\usepackage{bm}
\usepackage{amsmath}
\usepackage{graphicx}
\usepackage{amsmath}
\usepackage{float}
\usepackage[export]{adjustbox}
\usepackage{subcaption}
\usepackage{makecell}
\usepackage{titling}
\usepackage{natbib}

\makeatletter

\let\c@table\c@figure
\makeatother

\title{Multi-Horizon Forecasting for Limit Order Books: \\ Novel Deep Learning Approaches and Hardware Acceleration using Intelligent Processing Units}

\author{
  Zihao Zhang, Stefan Zohren\\
  Oxford-Man Institute of Quantitative Finance,\\
  University of Oxford
}

\begin{document}

\maketitle

\begin{abstract}
 
We design multi-horizon forecasting models for limit order book (LOB) data by using deep learning techniques. Unlike standard structures where a single prediction is made, we adopt encoder-decoder models with sequence-to-sequence and Attention mechanisms to generate a forecasting path. %An encoder steps through input time steps to extract representative features and a decoder allows short-term predictions to be fed into next estimation through an autoregressive structure. 
Our methods achieve comparable performance to state-of-art algorithms at short prediction horizons. Importantly, they outperform when generating predictions over long horizons by leveraging the multi-horizon setup. Given that encoder-decoder models rely on recurrent neural layers, they generally suffer from slow training processes. To remedy this, we experiment with utilising novel hardware, so-called Intelligent Processing Units (IPUs) produced by Graphcore. IPUs are specifically designed for machine intelligence workload with the aim to speed up the computation process. We show that in our setup this leads to significantly faster training times when compared to training models with GPUs. 

\end{abstract}

%We concatenate features from different indices to form an input and use neural networks to output an allocation distribution. 
%%%%%%%%%%%%%%%%%%%%%%%%%%

\section{Introduction}
\label{introduction}

Limit order books (LOBs), as the canonical example of high-frequency financial microstructure data, have received tremendous popularity in recent academic studies. 
At any time stamp, a LOB is a record of all outstanding limit orders (passive orders) for a financial instrument at an exchange. It is sorted into different levels based on the prices of the submitted orders. The LOB has two sides, representing buy and sell orders -- also referred to as bid and ask. Each level of a LOB indicates the total available volume (number of shares) at the price of that level. Those detailed records of price and volume information provide us with a picture of the short-term supply and demand relationship. From this, we can compute quantities such as order imbalances \citep{chordia2002order}, which help to understand the dynamics of high-frequency microstructure data. 

Recent works by \cite{tsantekidis2017forecasting, tran2018temporal, zhang2019deeplob, briola2020deep} demonstrate that LOBs have strong predictive to forecast price moves at short time intervals. Their findings have inspired a range of extensions including high-frequency trading models \citep{briola2021deep}. However, the aforementioned works are formulated as standard supervised learning tasks where a price at a single future point in time is predicted. Given that financial time-series are notoriously stochastic with a low signal-to-noise ratio \citep{gould2013limit}, a single prediction imposes limitations to describe the future evolution of market movements. Naturally, multi-horizon forecasting (predicting multiple steps into future) is desirable, since we can obtain a forecasting path which can be used for trading decision making or risk management.

In this work, we design multi-horizon forecasting models for LOB data with deep learning techniques \citep{Goodfellow-et-al-2016}. Inspired by the machine translation problems \citep{bahdanau2014neural} from Natural language processing (NLP), we apply sequence-to-sequence (Seq2Seq) \citep{sutskever2014sequence, cho2014learning} and Attention \citep{luong2015effective} models to generate multi-horizon forecasts. We adopt the deep network architecture from \cite{zhang2019deeplob} and engineer the output layer to produce a forecasting path. 
We test our method on the popular publicly available LOB dataset (FI-2010 \cite{ntakaris2018benchmark}) and one year order book data from the London Stock Exchange (LSE). The experiments show that our model delivers competitive results when compared to state-of-the-art models for single step, short horizon forecasts. Furthermore, in our setting a single network is capable of predicting multi-steps into future, avoiding the limitations of a single point estimation. Interstingly, our method delivers superior results for predicting over long horizons as short-term predictions contribute to future estimation through an autoregressive structure. 

The dominant Seq2Seq and Attention models are based on complex recurrent neural layers that include an encoder and a decoder. Such recurrent structures lead to substantially slow training processes even when employing GPUs for acceleration. This often poses challenges which need to be overcome. The work of \cite{vaswani2017attention}, for example, proposes Transformers to allow parallel training of attention mechanisms by using fully connected layers. 
In this work, we utilise a different form of hardware acceleration,  so-called Intelligence Processing Units (IPUs). IPUs developed by Graphcore \citep{graphcore} are a novel massively parallel processor. They can be used to accelerate the training process, offering an alternative solution to deal with this bottleneck. 
We compare the computation efficiency between GPUs and IPUs in our setting by benchmarking with a wide variety of state-of-the-art network architectures. The results indicate that IPUs are multiple times faster than GPUs. This significant improvement in computation is not necessarily restricted to the training process but could also lead to speedups in a wide range of applications within existing algorithms, for example, to reduce latency in market-making strategies. 

The remainder of this paper is structured as follows: In Section~\ref{literature}, we include a literature review to discuss the development of deep learning algorithms on LOBs and review multi-horizon forecasting models. Section~\ref{ipu} gives a short introduction to IPUs and Section~\ref{model} discusses our proposed network architectures. We then describe our experiments and present the results in Section~\ref{experiment}. We conclude in Section~\ref{conclusion} by summarising our findings and proposing potential research problems.

\section{Literature Review}
\label{literature}

Deep learning models have been heavily used for prediction tasks on LOB data, where \cite{tsantekidis2017forecasting, tsantekidis2017using, sirignano2019universal, zhang2019deeplob} helped to build the foundation in this area. Subsequently, a wide range of extensions have been proposed to improve predictive performance, including Bayesian deep networks \citep{zhang2018bdlob}, Quantile regression \citep{zhang2019extending}, Transformers \citep{wallbridge2020transformers} and usages of more granular market by order data \citep{zhang2021deep}. In addition, LOB data has been studied in the context of reinforcement learning \citep{wei2019model}, market-making \citep{sadighian2019deep}, cryptocurries \citep{jha2020deep}, forecasting quoted depth \citep{libman2021forecasting} and portfolio optimisation \citep{sangadiev2020deepfolio}. However, to the best of our knowledge, we have not found any existing work that studys multi-horizon forecasts for LOB data and we aim to fill this gap in the literature.

In terms of the multi-horizon forecasting models, \cite{taieb2010multiple, marcellino2006comparison} introduced traditional econometric approaches. In this work, we focus on recent deep learning techniques, mainly Seq2Seq \citep{sutskever2014sequence, cho2014learning} and attention-based \citep{luong2015effective, fan2019multi} approaches. A typical Seq2Seq model contains an encoder to summarise past time-series information and a decoder to combine hidden states with future known inputs to generate predictions. However, the Seq2Seq model only utilises the last hidden state from an encoder to make estimations, thus making it incapable of processing inputs with long sequences. Attention was proposed to assign a proper weight to each hidden state from the encoder to solve this limitation. We study both methods on LOB data and adapt the network architecture in \cite{zhang2019deeplob} to propose an end-to-end framework for generating multi-step predictions. 

Despite the popularity of Seq2Seq and Attention models, the recurrent nature of their structure imposes bottlenecks for training. This potentially limits their use cases on high-frequency microstructure data as modern electronic exchanges can generate billions of observations in a single day, making the training of such models on large and complex LOB datasets infeasible even with multiple GPUs. Here we experiment with IPUs for hardware acceleration. Graphcore introduced their IPUs as a novel massively parallel processor for training deep learning models \citep{ipu}. The work in \cite{jia2019dissecting} conducted a thorough investigation between GPUs and IPUs, observing massive speed-ups in training. We test the computational power of GPUs and IPUs on the state-of-art network architectures for LOB data and our findings are in line with \cite{jia2019dissecting}. We utilise Graphcore's Poplar graph framework software \citep{poplar} which allows for seamless integration of IPUs with TensorFlow, Keras \citep{abadi2016tensorflow} and PyTorch \citep{paszke2019pytorch}, requiring minimum emendation from existing repositories written for GPUs.

\section{Intelligence Processing Unit (IPU)}
\label{ipu}

Graphcore has designed IPUs specifically for machine intelligence problems and some of the computing architectures differ radically from common hardware such as CPUs and GPUs. In this section, we present a brief introduction to IPUs and discuss some differences among these architectures. For a complete and in-depth comparison, interested readers are referred to \cite{jia2019dissecting}.

\begin{figure}[!t]
\centering
\includegraphics[width=5.5in, height=3.5in]{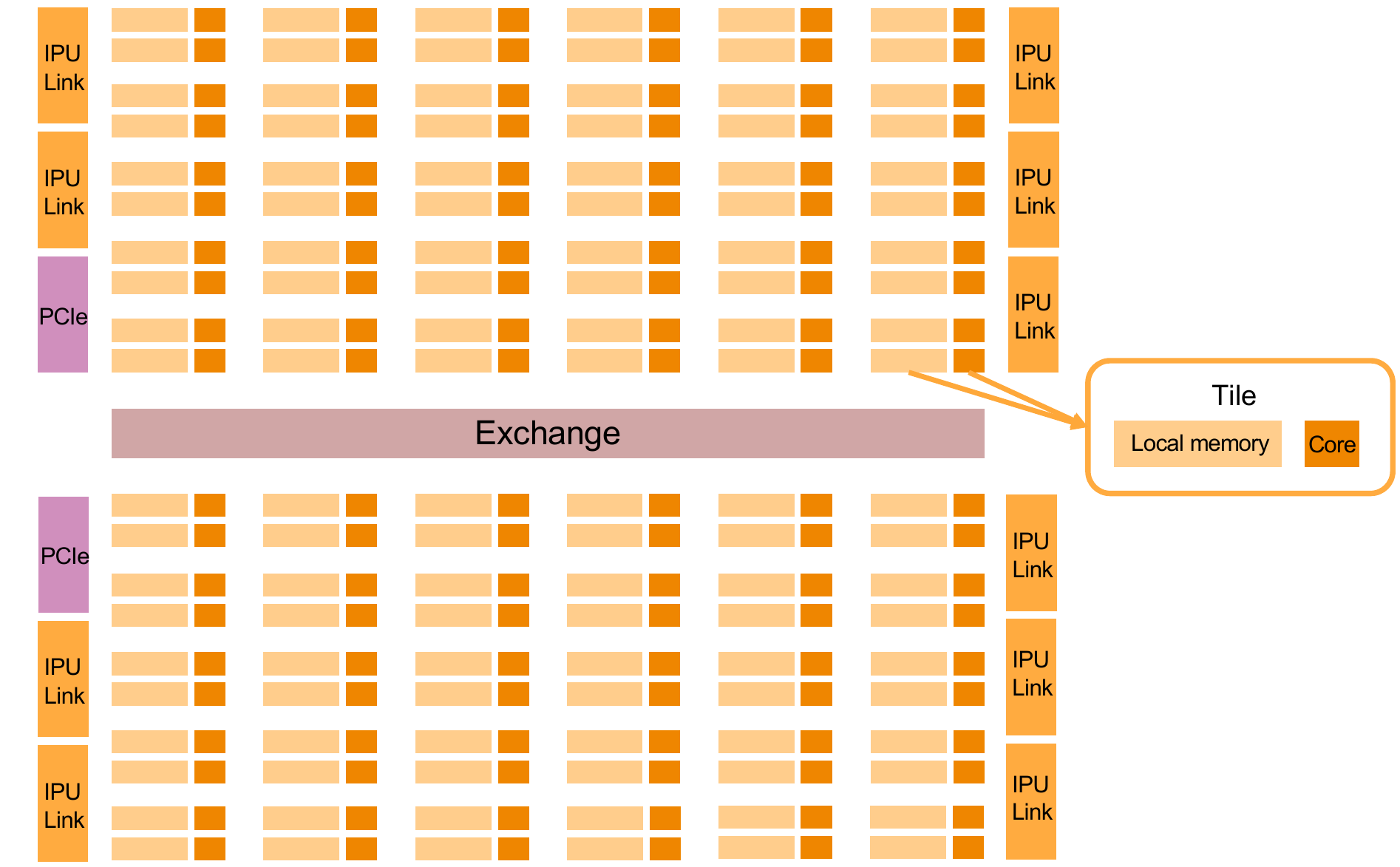}
\caption{Simplified illustration of an IPU processor. Each processor has four main components: The tile, exchange, link and PCIe. }
\label{fig:ipu}
\end{figure}

The cornerstone of an IPU-based system is the IPU processor with the aim of achieving efficient execution of fine-grained operations across a relatively large number of parallel threads. In general, each IPU processor contains four components: IPU-tile, IPU-exchange, IPU-link and PCIe. For each processor, there are 1216 tiles and each tile consists of one computing core and 256 kB of local memory. These tiles are interconnected by the IPU-exchange which allows for low-latency and high-bandwidth communication. In addition, each IPU contains ten IPU-link interfaces, which is a Graphcore proprietary interconnect that enables low latency, high-throughput communication between IPU processors. IPU-links enable transfers between remote tiles as efficient as between local tiles and they are key to the IPU's computational scalability. Besides that, each IPU contains two PCIe links for communication with CPU-based hosts. We illustrate the IPU architecture with a simplified diagram in Figure~\ref{fig:ipu}.

The architecture of IPUs differs significantly from CPUs and GPUs that are commonly used for training machine learning algorithms. In general, CPUs excel at single-thread performance as they offer complex cores in relatively small counts. However, even with the vectorisation of data, CPUs are incomparable with GPUs in aggregate floating-point arithmetic on large and complex workloads.  GPUs, on the other hand, have architecturally simpler cores than CPUs but do not offer branch speculation or hardware prefetching. The typical arrangment of GPUs are grouped into clusters, so all cores in a cluster execute the same instruction at any point in time. Because of this architecture, GPUs are proficient at regular, dense, numerical, data-flow-dominated workloads and tend to be more energy efficient than CPUs.

The IPU's approach of accelerating computation is through shared memory, which is distinct from the other hardware. An IPU offers small and distributed memories that are locally coupled to each other, therefore, IPU cores pay no penalty when their control flows diverge or when the addresses of their memory accesses diverge. Such a structure allows cores to access data from local memory at a fixed cost that is independent of access patterns, making IPUs more efficient than GPUs when executing workloads with irregular or random data access patterns as long as the workloads can be fitted in IPU memory. 

In this work, we compare the training efficiency between GPUs and IPUs for state-of-art deep networks on LOB data and the results indeed suggest that IPUs deliver superior computation performance.

\section{Multi-Horizon Forecasting Models}
\label{model}

This section introduces deep learning architectures for multi-horizon forecasting models, in particular Seq2Seq and Attention models. In essence, both of these architectures consist of three components: an encoder, a context vector and a decoder. We write a given input of any time-series as $\bm{x}_{1:T} = (x_1, x_2, \cdots, x_T) \in \mathbb{R}^{T \times m}$,  where $T$ represents the length of the input sequence and $x_t$ indicates $m$ features at each time stamp $t$. Similarly, a typical output is $\bm{y}_{1:k} = (y_1, y_2, \cdots, y_k) \in \mathbb{R}^{k \times n}$ where $k$ is the furthest prediction point, and it is essentially a multi-input and multi-output setup. 

An encoder steps through the input time steps to extract meaningful features. The resulting context vector encapsulates the resulting sequence into a vector for integrating information. Finally, a decoder reads from the context vector and steps through the output time step to generate multi-step predictions. The fundamental difference between the Seq2Seq and Attention model is the construction of the context vector. The Seq2Seq model only takes the last hidden state from the encoder to form the context vector, whereas the Attention model utilises the information from all hidden states in the encoder.

\subsection{Sequence to Sequence Learning (Seq2Seq)}

\begin{figure}[!t]
\centering
\includegraphics[width=4in, height=2.2in]{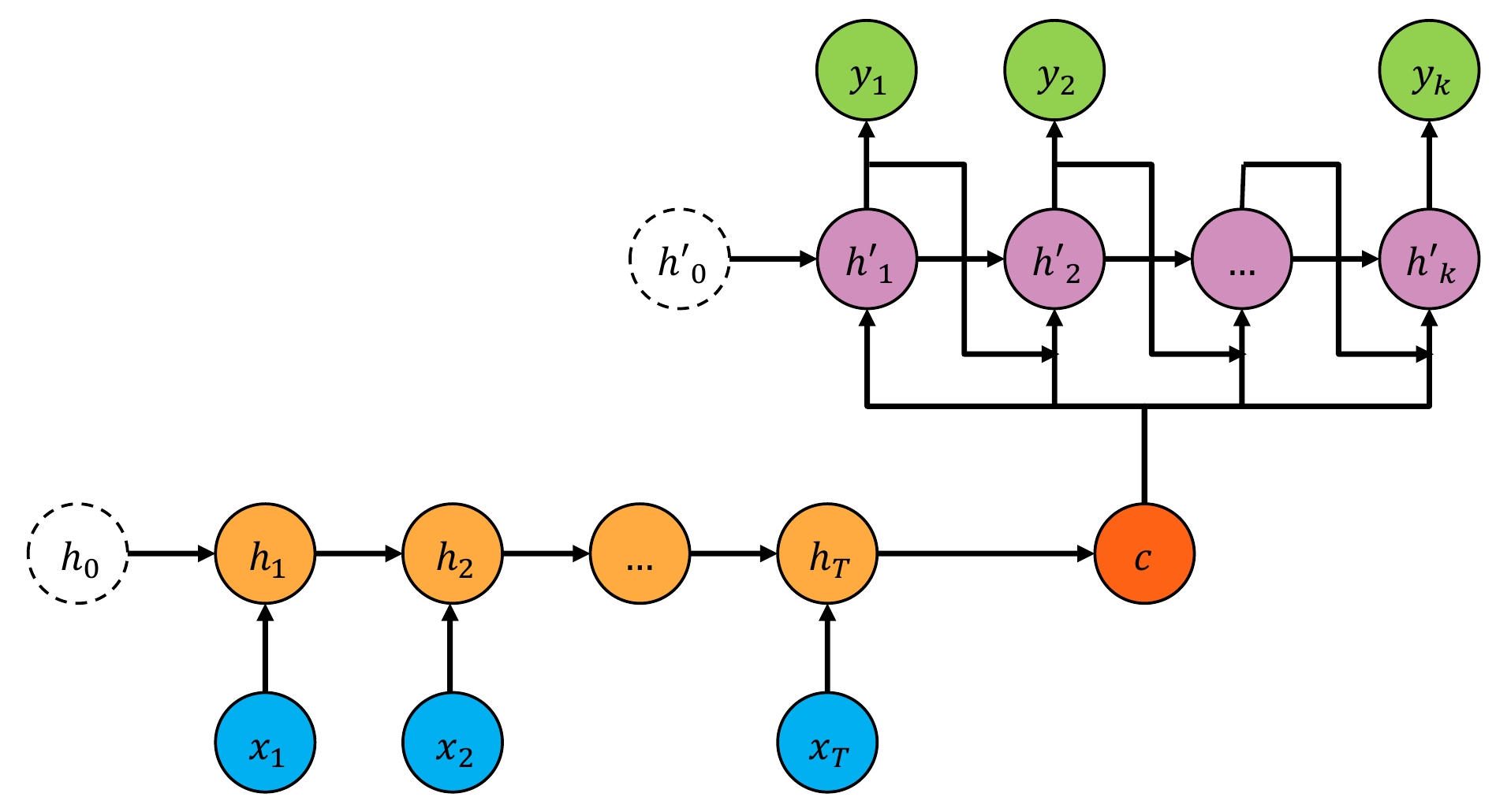}
\caption{A schematic depiction of the Seq2Seq network architecture. }
\label{fig:seq2seq}
\end{figure}

In this work, we employ the Seq2Seq architecture in \cite{cho2014learning} in the context of multi-horizon forecasting models for LOBs. Overall, the encoder contains a recurrent neural network (RNN) that operates on a given input $\bm{x}_{1:T} = (x_1, x_2, \cdots, x_T)$. At each time step $t$ of the encoder, the hidden state $\bm{h}_t$ is
\begin{equation}
\text{Encoder:} \quad \bm{h}_t = f(\bm{h}_{t-1}, x_t),
\end{equation}
where $f$ is a nonlinear function. The choice of $f$ can vary, ranging from a simple logistic sigmoid function to a more complex LSTM \citep{hochreiter1997long}. The encoder reads through a given input and the last hidden state summarises the whole sequence. The last hidden state $\bm{c}$ is the ``bridge'' between the encoder and decoder, also known as the context vector. At time step $t$ of the decoder, the hidden state $\bm{h}'_t$ is 
\begin{equation}
\text{Decoder:} \quad \bm{h}'_t = f(\bm{h}'_{t-1}, y_{t-1}, \bm{c}),
\end{equation}
and the distribution for output $y_{t}$ is
\begin{equation}
P(y_{t} | y_{t-1}, y_{t-2}, \cdots, y_{1}, \bm{c}) = g(\bm{h}'_{t}, \bm{c}).
\end{equation}
Here $f$ and $g$ are nonlinear functions and $g$ needs to produce valid probabilities, which in our case is done through a softmax activation function. Figure~\ref{fig:seq2seq} illustrates the structure of a standard Seq2Seq network.
Seq2Seq models work well for inputs with small sequences, but suffers when the length of the sequence increases as it is difficult to summarise the entire input into a single hidden state represented by the context vector. Models tend to forget the earlier parts of the input and results often deteriorate as the size of the sequence increases.

\subsection{Attention}

The Attention model \citep{luong2015effective} is an evolution of the Seq2Seq model, developed in order to deal with inputs of long sequences. The core idea is to allow the decoder to selectively access hidden states of the encoder during decoding. We can build a different context vector for every time step of the decoder as a function of the previous hidden state and of all the hidden states in the encoder. Similar to the Seq2Seq model, the hidden state $\bm{h}_t$ of the encoder is
\begin{equation}
\text{Encoder:} \quad \bm{h}_t = f(\bm{h}_{t-1}, x_t),
\end{equation}
where $f$ is a nonlinear function. For the context vector $\bm{c}_t$ at time stamp $t$ of the decoder, one defines
\begin{equation}
  \begin{alignedat}{2}
    & \text{Context vector:} &&\bm{c}_t = \sum_{i=1}^{T} \alpha_{t,i} h_i, \\
    & \text{Attention weight:} \quad &&\alpha_{t,i} = \frac{exp(e(\bm{h}'_{t-1}, \bm{h}_i))}{\sum_{j=1}^Texp(e(\bm{h}'_{t-1}, \bm{h}_j))},
  \end{alignedat}
\end{equation}
where $e(\bm{h}'_{t-1}, \bm{h}_i)$ is called the score derived from the previous hidden state $\bm{h}'_{t-1} $ of the decoder and the hidden state $\bm{h}_i$ of the encoder. In \cite{luong2015effective}, there are three alternatives to calculate the score
\begin{equation}
  e(\bm{h}'_{t-1}, \bm{h}_i) =
    \begin{cases}
      \bm{h}_i^{\text{T}} \bm{h}'_{t-1} & \text{dot},\\
      \bm{h}_i^{\text{T}} \bm{W}_a \bm{h}'_{t-1} & \text{general},\\
      tanh(\bm{W}_a [\bm{h}_i^{\text{T}} ; \bm{h}'_{t-1}]) & \text{concatenate}.
    \end{cases}       
\end{equation}

\begin{figure}[!t]
\centering
\includegraphics[width=4in, height=2.5in]{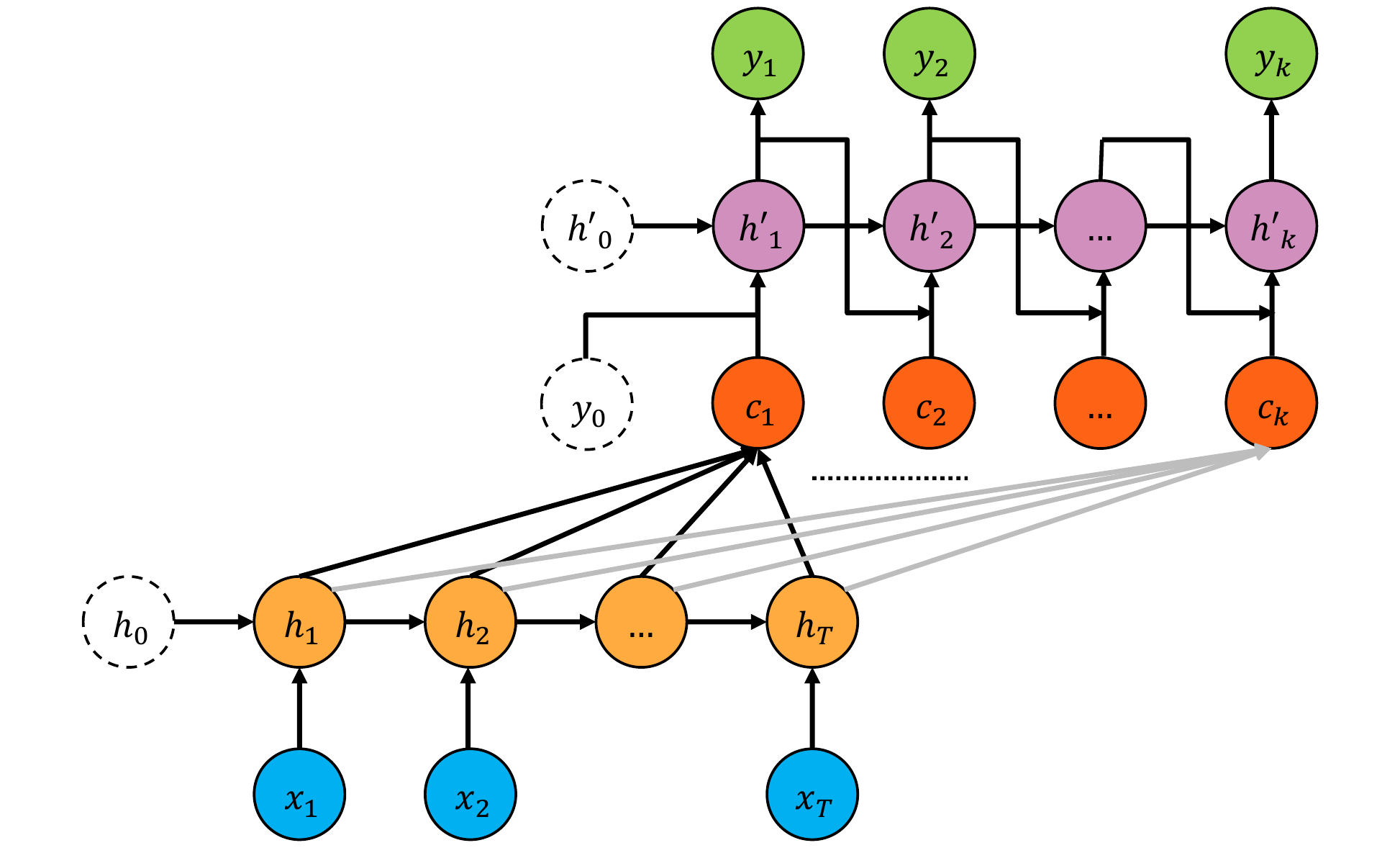}
\caption{An schematic depiction of the Attention mechanism. }
\label{fig:attention}
\end{figure}

We can then pass the context vector $\bm{c}_t$ to the decoder to calculate the probability distribution of the next possible output
\begin{equation}
\begin{split}
\text{Decoder:} \quad \bm{h}'_t &= f(\bm{h}'_{t-1}, y_{t-1}, \bm{c}_t), \\
P(y_{t} | y_{t-1}, y_{t-2}, \cdots, y_{1}, \bm{c}_t) &= g(\bm{h}'_{t}, \bm{c}_t),
\end{split}
\end{equation}
where $\bm{h}'_{t}$ is the hidden state of the decoder at time $t$ and $g$ is a softmax activation function. We illustrate the Attention mechanism in Figure~\ref{fig:attention}.

\subsection{Network Architecture}

LOBs are complex dynamic objects of high dimensionality. Furthermore, LOB data, just like any other financial time-series, is notoriously non-stationary and of low signal-to-noise ratio \citep{gould2013limit}. Since the encoder reads through an input to extract meaningful information, we adapt the modern deep network (DeepLOB) designed specifically for limit order books in \cite{zhang2019deeplob} as the encoder, extracting representative features from raw LOB data.  Here we give a brief introduction to DeepLOB -- the exact network architecture can be found in \cite{zhang2019deeplob}.  Overall,  DeepLOB comprises three building blocks: a convolutional block with multiple convolutional layers (CNNs),  an Inception Module and a LSTM layer.  

%\textcolor{red}{
\paragraph{The convolutional block:} The usage of CNNs is to automate the process of feature extraction as the convolutional layer can directly deal with grid-like data structures. It possesses nice properties like smoothing and parameter sharing,  where the latter is important to deal with the low signal-to-noise ratio. Note that in the time dimension the CNN acts similar to an auto-regressive model.  The convolutional block processes the raw LOB data and extracts representative features from different order book levels. 
%}

%\textcolor{red}{
\paragraph{Inception Module:} The resulting feature maps from the convolutional block are passed into the Inception Module which consists of multiple convolutional layers in parallel -- each with a different kernel size -- to infer local interactions over different time horizons.  Instead of adding layers vertically to a network, the Inception Module expands the model horizontally so we can capture different dynamic behaviours of the time-series by wrapping several convolutions together.
%}

\begin{figure}[!t]
\centering
\includegraphics[width=3.8in, height=1.9in]{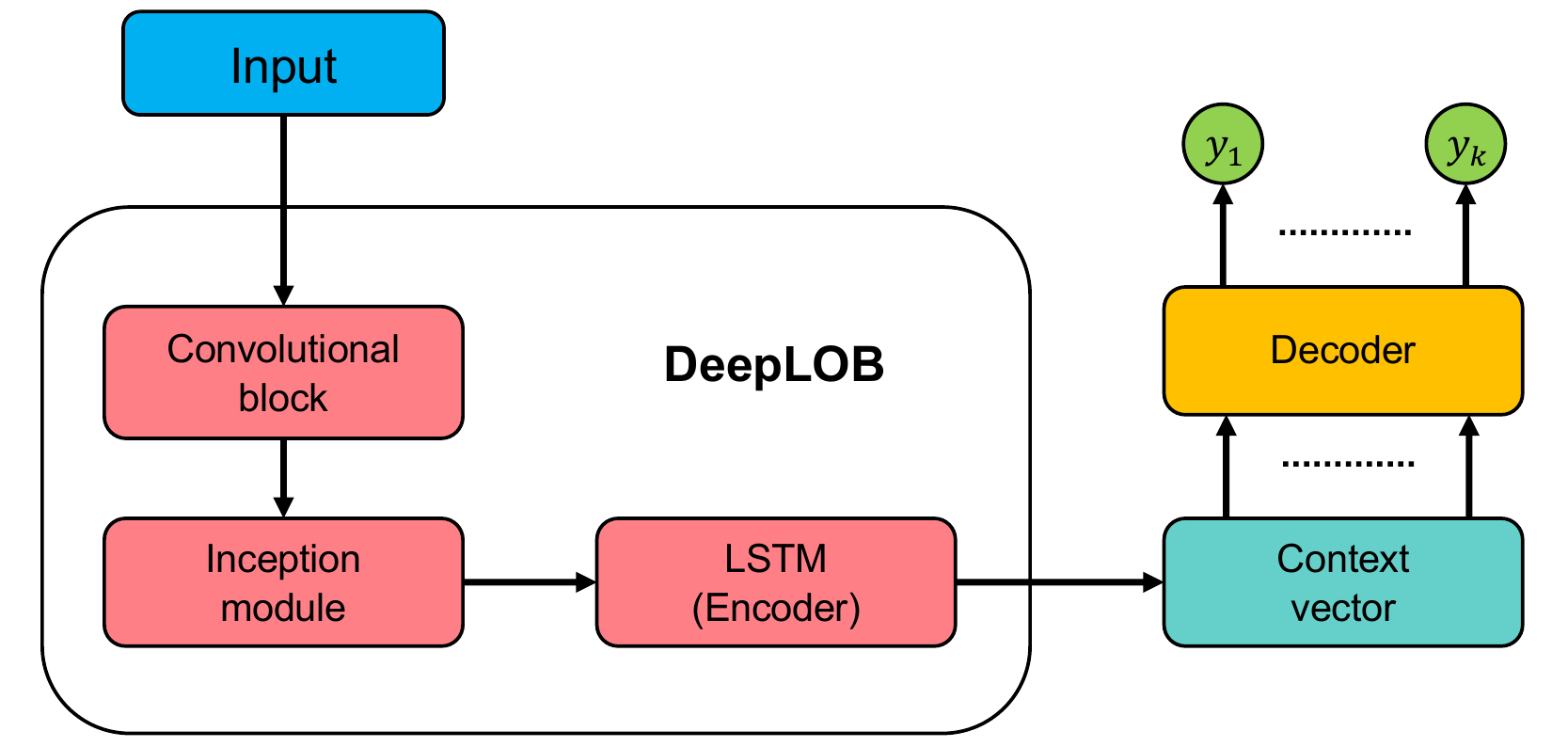}
\caption{Model architecture schematic: A combination of DeepLOB and an encoder-decoder structure. }
\label{fig:deeplob}
\end{figure}

%\textcolor{red}{
\paragraph{LSTM:} Lastly, a LSTM is used to capture the longer temporal behaviour in the resulting features and it also servers as the encoder of our multi-horizon forecasting model. 
We illustrate the resulting model architecture in Figure~\ref{fig:deeplob}. 

For the decoder, we experiment with both Seq2Seq and Attention models to generate multi-horizon forecasts. An interesting byproduct of using Attention models is that the attention weights can be used to understand the importance of input features. 
%}

\section{Experiments}
\label{experiment}

\subsection{Descriptions of Datasets}

The FI-2010 dataset \citep{ntakaris2018benchmark} is the first publicly available benchmark dataset of high-frequency limit order book data. Many previous algorithms are tested on this dataset and we use it to build a fair comparison between our method and state-of-art algorithms. The FI-2010 dataset consists of LOB updates for five stocks from the Nasdaq Nordic stock exchange for a time period of 10 consecutive days. It contains 10 levels for both ask and bid of a LOB and each level has information for price and available volume at that price level. A classification setup is formulated in which we have three classes of labels: pricing going up, staying stationary or going down. Also, it studied five prediction horizons at $k = (10, 20, 30, 50, 100)$ in ``tick time'', i.e. consecutive LOB updates. As argued in multiple place, tick time, similar to volume time, is a natural time to consider for financial instruments. 

The exact procedure of data normalisation and label formation can be found in \cite{ntakaris2018benchmark} and we follow the setting originated from the works \citep{tsantekidis2017forecasting, tsantekidis2018using} to evaluate our network architecture, in which the first 7 days are used as the train data and the last 3 days as the test data. We split the last 20\% observations from the train set as the validation set to optimise hyperparameters. Each input contains the most recent 50 updates and each update includes information for both ask and bid of a LOB, therefore, a single input $\bm{x}_{1:T} \in \mathbb{R}^{T \times m}$ has the dimension of $(50, 40)$. We feed inputs to the designed network where the encoder extracts representative features and the decoder generates multi-horizon forecasts. In this work, our model can directly forecast all 5 prediction horizons in contrast to standard methods where a separate model is needed for each prediction horizon.  

A major drawback of the FI-2010 dataset is that it is limited in scope and size. To address those shortcomings we further test our methods by using LOB data for five highly liquid stocks --  Lloyds (LLOY),  Barclays (BARC),  Tesco (TSCO),  BT and Vodafone (VOD) -- for the entire year of 2018 from the London Stock Exchange (LSE).  The LSE dataset contains bid and ask information for an order book up to ten levels and, for each trading day, we take the data between 08:30:00 and 16:00:00, restricting ourselves to liquid continuous trading hours, excluding any auctions. Overall, we have more than 169 million samples in our dataset and we take the first 6 months as training data, the next 3 months as validation data and the last 3 months as testing data.

%\textcolor{red}{
The works of \cite{sirignano2019universal, zhang2019deeplob} show that deep networks trained on LOB data can be applied to predict stocks that are not even part of the training set -- this is sometimes referred to as transfer learning. %\textbf{(transfer learning)}. 
We investigate this observation on our multi-horizon forecasting models to further verify the generalisation and robustness of our methods. To test this universality, we directly apply our models to 20 more stocks from the London Stock Exchange. These 20 stocks have the same testing period as the original 5 stocks that are used to calibrate model weights, and the label classes are roughly balanced for each instrument. 
%}

The categorical cross-entropy loss is our objective function, and we use four evaluation metrics: Accuracy (Acc.), Precision (Pre.), Recall (Re.) and F1-score. Since the labels in FI-2010 dataset are not well balanced, \cite{ntakaris2018benchmark} suggested to focus on F1 score as the main evaluation metric and Kolmogorov-Smirnov \citep{massey1951kolmogorov} tests are used to check how results are statistically different. The code is available at GitHub \footnote{\url{https://github.com/zcakhaa/Multi-Horizon-Forecasting-for-Limit-Order-Books} \label{code}}.

\subsection{Experimental Results on the FI-2010 Dataset}

As described above, we adapt DeepLOB \citep{zhang2019deeplob} as our encoder and further details of the architecture and hyperparameters can be found in our GitHub repository. In terms of the decoder, we use a single LSTM with 64 units for both Seq2Seq and Attention, denoted as DeepLOB-Seq2Seq and DeepLOB-Attention respectively. We include a wide variety of benchmark algorithms in the experiment, including a support vector machine (SVM \citep{tsantekidis2017using}), a multi-layer perceptron (MLP \citep{tsantekidis2017using}), a convolutional network (CNN-I \citep{tsantekidis2017forecasting}), a LSTM (\citep{tsantekidis2017using}), a variant convolutional network (CNN-II \citep{tsantekidis2018using}), as well as an Attention-augmented-Bilinear-Network with one hidden layer (B(TABL) \citep{tran2018temporal}) and two hidden layers (C(TABL) \citep{tran2018temporal}). Note that these benchmark algorithms produce a single-point estimation and the authors did not test on all prediction horizons available for the FI-2010 dataset.

\begin{table}[!p]
\caption{Experiment results for the FI-2010 dataset.}
\label{table:benchmark}
\centering
\begin{tabular}{l|llll}
\toprule
\textbf{Model}    & \textbf{Accuracy} \% & \textbf{Precision} \% & \textbf{Recall} \%& \textbf{F1} \% \\
\midrule
\multicolumn{5}{c}{\textbf{Prediction Horizon k = 10}} \\
\midrule
SVM \citep{tsantekidis2017using}      &-          &39.62      &44.92      &35.88    \\
MLP \citep{tsantekidis2017using}      &-          &47.81      &60.78      &48.27    \\
CNN-I \citep{tsantekidis2017forecasting}&-          &50.98      &65.54      &55.21    \\
LSTM \citep{tsantekidis2017using}     &-          &60.77      &75.92      &66.33    \\
CNN-II \citep{tsantekidis2018using}   &-          &56.00      &45.00      &44.00    \\
B(TABL) \citep{tran2018temporal}      &78.91      &68.04      &71.21      &69.20    \\
C(TABL) \citep{tran2018temporal}      &84.70      &76.95      &78.44      &77.63    \\
DeepLOB  \citep{zhang2019deeplob}                       &84.47      &84.00      &84.47      &83.40\\
\midrule
%DeepLOB-Direct &83.03      &82.42      &83.03      &81.61 \\
DeepLOB-Seq2Seq &82.58      &81.65      &82.58      &81.51 \\
DeepLOB-Attention &83.28      &82.50      &83.28      &82.37 \\
\midrule
\multicolumn{5}{c}{\textbf{Prediction Horizon k = 20}} \\
\midrule
SVM \citep{tsantekidis2017using}      &-          &45.08      &47.77      &43.20    \\
MLP \citep{tsantekidis2017using}      &-          &51.33      &65.20      &51.12    \\
CNN-I \citep{tsantekidis2017forecasting}&-         &54.79      &67.38      &59.17    \\
LSTM \citep{tsantekidis2017using}     &-          &59.60      &70.52      &62.37    \\
CNN-II \citep{tsantekidis2018using}   &-          &-      &-      &-    \\
B(TABL) \citep{tran2018temporal}      &70.80      &63.14      &62.25      &62.22    \\
C(TABL) \citep{tran2018temporal}      &73.74      &67.18      &66.94      &66.93    \\
DeepLOB  \citep{zhang2019deeplob}                                                 &74.85      &74.06      &74.85      &72.82    \\
\midrule
%DeepLOB-Direct &74.96      &73.99      &74.96      &73.08 \\
DeepLOB-Seq2Seq &74.38      &73.12      &74.38      &72.99 \\
DeepLOB-Attention &75.25      &74.31      &75.25      &73.73 \\
\midrule
\multicolumn{5}{c}{\textbf{Prediction Horizon k = 30}} \\
\midrule
CNN-I \citep{tsantekidis2017forecasting} & 67.98     & 66.52    &67.98      &65.72    \\
DeepLOB  \citep{zhang2019deeplob}                                                 & 76.36     & 76.00    & 76.36     & 75.33   \\
\midrule
%DeepLOB-Direct &76.87      &76.62      &76.87      &75.82\\
DeepLOB-Seq2Seq &76.41      &75.86      &76.41      &75.75 \\
DeepLOB-Attention &77.59      &77.32      &77.59      &76.94 \\
\midrule
\multicolumn{5}{c}{\textbf{Prediction Horizon k = 50}} \\
\midrule
SVM \citep{tsantekidis2017using}      &-          &46.05      &60.30      &49.42    \\
MLP \citep{tsantekidis2017using}      &-          &55.21      &67.14      &55.95    \\
CNN-I \citep{tsantekidis2017forecasting}&-          &55.58      &67.12      &59.44    \\
LSTM \citep{tsantekidis2017using}     &-          &60.03      &68.58      &61.43    \\
CNN-II \citep{tsantekidis2018using}   &-          &56.00      &47.00      &47.00    \\
B(TABL) \citep{tran2018temporal}      &75.58      &74.58      &73.09      &73.64    \\
C(TABL) \citep{tran2018temporal}      &79.87      &79.05      &77.04      &78.44    \\
BL-GAM-RHN-7 \citep{luo2019recurrent}&82.02      &81.45      &80.43      &80.88   \\
DeepLOB \citep{zhang2019deeplob}                                                  &80.51      &80.38      &80.51      &80.35   \\
\midrule
%DeepLOB-Direct &78.73      &78.54      &78.73      &78.40 \\
DeepLOB-Seq2Seq &78.10      &77.96      &78.10      &77.99 \\
DeepLOB-Attention &79.49      &79.51      &79.49      &79.38 \\
\midrule
\multicolumn{5}{c}{\textbf{Prediction Horizon k = 100}} \\
\midrule
CNN-I \citep{tsantekidis2017forecasting} & 64.87     &65.51     & 64.87     & 65.05   \\
DeepLOB  \citep{zhang2019deeplob}                                                 & 76.72     & 76.85    &  76.72    & 76.76   \\
\midrule
%DeepLOB-Direct &79.71      &79.66      &79.70      &79.66 \\
DeepLOB-Seq2Seq &79.09      &79.31      &79.09    &79.16\\
DeepLOB-Attention &81.45      &81.62      &81.45      &81.49 \\
\bottomrule
\end{tabular}
\end{table}

Table~\ref{table:benchmark} summarises the results for all models studied at each prediction horizon and all results are statistically different in terms of the Kolmogorov-Smirnov test. 
We observe that both DeepLOB-Seq2Seq and DeepLOB-Attention deliver comparable results to the state-of-art benchmark algorithms. Specifically, with shorter prediction horizons ($k=10, 20, 30, 50$), the performance gap between DeepLOB and our methods is very small. However, both our new multi-horizon forecasting models achieve superior results for predicting a long horizon ($k=100$), with DeepLOB-Attention being the best. The architecture of a decoder allows short-term predictions to be fed into the next estimation and our results indicate that this autoregressive structure helps with longer prediction horizons through the iterative estimation procedure.  

\begin{figure}[!t]
\centering
\includegraphics[width=5.5in, height=1.1in]{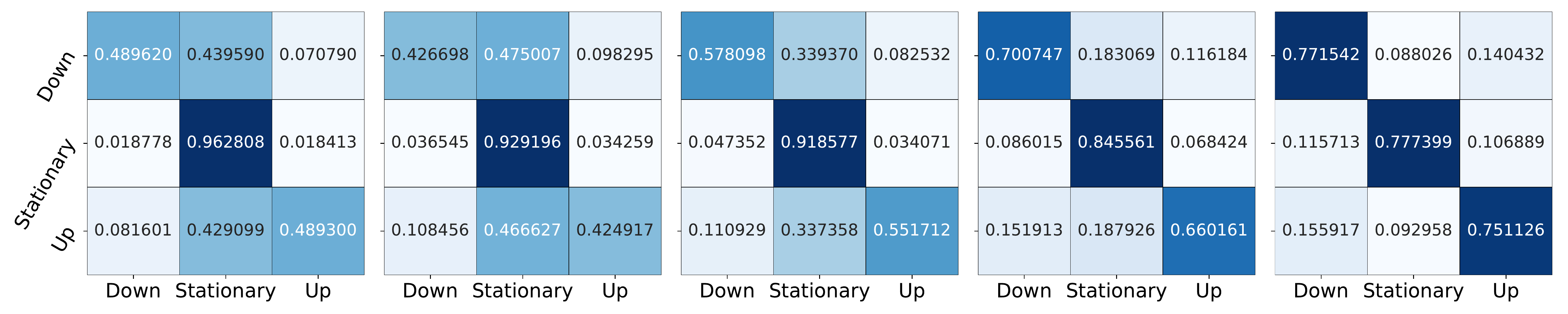}
\includegraphics[width=5.5in, height=1.1in]{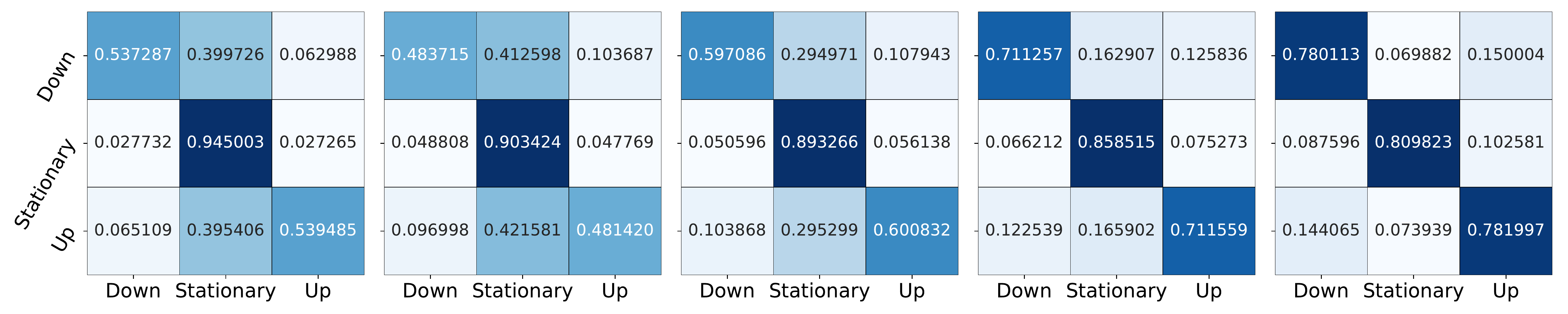}
\includegraphics[width=5.5in, height=1.1in]{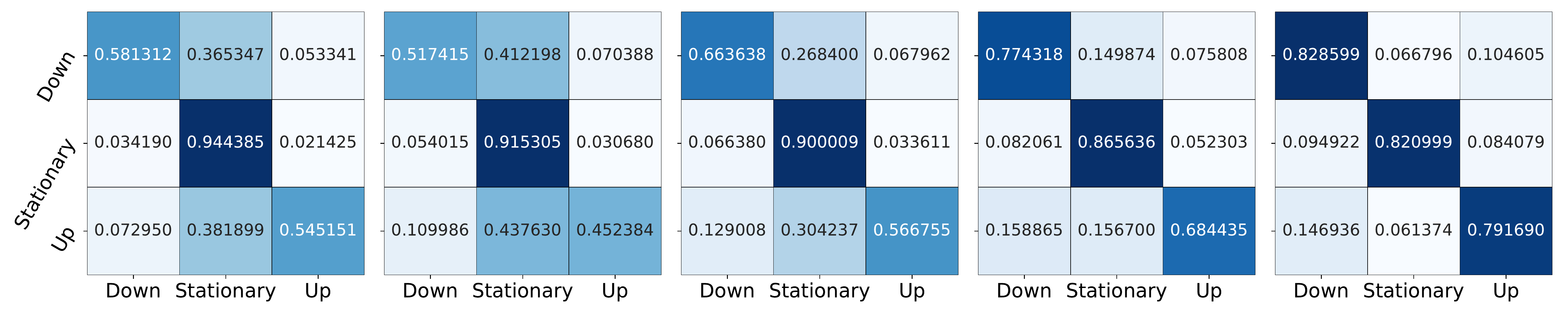}
\caption{Normalised confusion matrices for DeepLOB (\textbf{TOP}), DeepLOB-Seq2Seq (\textbf{Middle}) and DeepLOB-Attention (\textbf{Bottom}). From the left to right, the prediction horizon ($k$) equals to 10, 20, 30, 50 and 100. }
\label{fig:confusion_matrix}
\end{figure}

The normalised confusion matrices for DeepLOB and our methods are presented in Figure~\ref{fig:confusion_matrix}. We use these plots to understand how models perform at predicting each class label. In general, all three methods achieve better accuracy for predicting stationary labels at short prediction horizons but the performance deteriorates as the horizon increases. Interestingly, all three models are getting better at predicting up and down labels with an increase in prediction horizon. Price movements are more significant helping to distinguish up or down labels from stationary labels. As a result, the models can do better at detecting larger price moves.

\begin{figure}[!t]
\centering
\includegraphics[width=5.5in, height=1.5in]{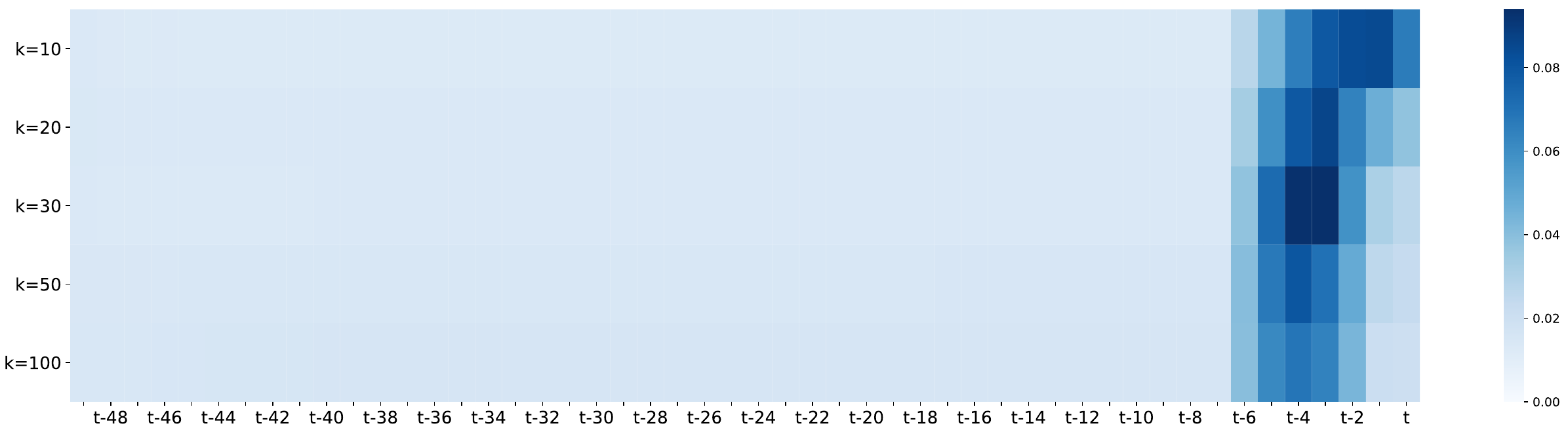}
\caption{Attention weights for features from the encoder of DeepLOB-Attention. }
\label{fig:attention_weights}
\end{figure}

When comparing our two networks with each other, we observe that DeepLOB-Attention delivers better results than DeepLOB-Seq2Seq. However, it seems that the Seq2Seq model does not suffer from long input sequences of LOB data. One possible explanation is that the markets are sufficiently efficient so that the current LOB observation contains most available information and features distant from current time stamp contain minimal predictive information. 
To verify this statement, we use the attention weights from DeepLOB-Attention to understand feature importance at different time steps from the encoder. Figure~\ref{fig:attention_weights} shows the plot of attention weights. We observe that the weights are largest for most recent observations with any features further in the past being inactive. This helps to explain why the Seq2Seq model does not suffer from long input sequences as past information is less relevant in this case. 

\subsection{Experimental Results on the LSE Dataset}

In terms of the LSE dataset, we follow the same setup as in \cite{zhang2019deeplob, zhang2021deep}.  We test on three prediction horizons $(k = 20, 50, 100)$ and list the choices of label parameters ($\alpha$) for the class threshold in Table~\ref{tb:setting} in Appendix~\ref{train_lse}. The label parameter ($\alpha$) is chosen for each instrument to have a balanced training set and the distribution of different class labels. More detail is provided in in Appendix~\ref{train_lse}. As before, we compare our methods with the following algorithms: a linear model (LM),  the multilayer perception (MLP) in  \cite{zhang2021deep},  LSTM \citep{sirignano2019universal},  CNN-I \citep{tsantekidis2017forecasting}, and DeepLOB \citep{zhang2019deeplob}.

Table~\ref{table:lse_results} presents the out-of-sample results obtained on the LSE dataset for the original 5 stocks in the training set. All results are statistically significant even though the absolute differences in evaluation metrics seem small.  Since the testing data is slightly unbalanced,  we focus on the F1-score as the main evaluation metric.  We observe that DeepLOB-Seq2Seq and DeepLOB-Attention deliver similar results to DeepLOB at short prediction horizons but both models perform better when considering longer prediction horizons as the decoder allows short-term predictions to be fed into the next estimation through an autoregressive structure.  For the LSE dataset,   we have more than 46 million samples in the testing set,  which should ensure the generalisation and robustness of our methods.

\begin{table}[!t]
\caption{Experiment results for the LSE dataset.}
\label{table:lse_results}
\centering
\begin{tabular}{l|llll}
\toprule
\textbf{Model}    & \textbf{Accuracy} \% & \textbf{Precision} \% & \textbf{Recall} \%& \textbf{F1} \% \\
\midrule
\multicolumn{5}{c}{\textbf{Prediction Horizon k = 20}} \\
\midrule
LM        & 45.71       & 43.44        & 45.71     & 42.38 \\
MLP \citep{zhang2021deep}       & 50.06       & 50.04        & 50.06     & 46.89 \\
LSTM \citep{sirignano2019universal}      & 66.09       & 67.53        & 66.09     & 66.68 \\
CNN-I \citep{tsantekidis2017forecasting}       & 63.39       & 67.31        & 63.39     & 64.64 \\
DeepLOB \citep{zhang2019deeplob}    &68.73       & 68.16        & 68.73     & \textbf{68.40}\\
DeepLOB-Seq2Seq  &67.59      & 67.94        & 67.59     & 67.76\\
DeepLOB-Attention    & 67.94       & 68.26       & 67.94     & 68.09\\
\midrule
\multicolumn{5}{c}{\textbf{Prediction Horizon k = 50}} \\
\midrule
LM        & 46.97       & 44.34        & 46.97     & 41.13 \\
MLP \citep{zhang2021deep}       & 50.56       & 48.46        & 50.56     & 47.25 \\
LSTM \citep{sirignano2019universal}      & 64.49       & 64.88      & 64.49     & 64.65 \\
CNN-I \citep{tsantekidis2017forecasting}       & 64.77       & 62.55        & 64.77     & 63.26 \\
DeepLOB \citep{zhang2019deeplob}   & 65.38       & 64.37        & 65.38     & 64.79\\
DeepLOB-Seq2Seq  &64.61      & 65.85        & 64.61     & 65.16\\
DeepLOB-Attention    & 65.53      & 65.75        &65.53     & \textbf{65.63}\\
\midrule
\multicolumn{5}{c}{\textbf{Prediction Horizon k = 100}} \\
\midrule
LM         & 46.19       & 43.29        & 46.19     & 41.80 \\
MLP \citep{zhang2021deep}       & 48.36       & 47.39        & 48.36     & 43.66 \\
LSTM \citep{sirignano2019universal}      & 61.27       & 58.47        & 61.27     & 57.96 \\
CNN-I \citep{tsantekidis2017forecasting}       & 61.78       & 56.91        & 61.78     & 55.40 \\
DeepLOB \citep{zhang2019deeplob}   & 62.82       & 60.94    & 62.82     & 61.10\\
DeepLOB-Seq2Seq  &61.54      & 62.45        & 61.54     & 61.95\\
DeepLOB-Attention    & 62.42       & 62.50        & 62.42     & \textbf{62.46}\\
\bottomrule
\end{tabular}
\end{table}

%\textcolor{red}{
%\paragraph{Transfer Learning}
In terms of transfer learning, Table~\ref{table:lse_transfer} shows the results where we directly apply DeepLOB, DeepLOB-Seq2Seq and DeepLOB-Attention (trained on the previous 5 instruments) to 20 stocks that are not part of the training set. We study the out-of-sample performance for our methods in both timing and data stream sense to verify the robustness and generalisation ability. Overall, all three models achieve strong predictive results with DeepLOB-Attention demonstrating an additional edge in predictive performance. The detailed results on each instrument can be found at Table~\ref{table:lse_seq_tranfer} and~\ref{table:lse_att_tranfer} in Appendix~\ref{train_lse}.This observation aligns with the findings in \cite{sirignano2019universal, zhang2019deeplob} and suggests the existence of universal features that characterise the demand and supply relationship in high-frequency LOB data.
%}

\begin{table}[!t]
\caption{Experiment results for transfer learning.}
\label{table:lse_transfer}
\centering
\begin{tabular}{l|llll}
\toprule
\textbf{Model}    & \textbf{Accuracy} \% & \textbf{Precision} \% & \textbf{Recall} \%& \textbf{F1} \% \\
\midrule
\multicolumn{5}{c}{\textbf{Prediction Horizon k = 20}} \\
\midrule
DeepLOB \citep{zhang2019deeplob}      &65.05       & 64.96        & 65.05     &64.83\\
DeepLOB-Seq2Seq     &65.39       & 65.22        & 65.39     & 65.29\\
DeepLOB-Attention   &65.51      & 65.23        & 65.51     & 65.35\\
\midrule
\multicolumn{5}{c}{\textbf{Prediction Horizon k = 50}} \\
\midrule
DeepLOB \citep{zhang2019deeplob}    &62.26       & 61.16       &62.26     & 61.02\\
DeepLOB-Seq2Seq     &62.92       & 61.63       &62.92     & 61.91\\
DeepLOB-Attention   &62.91      & 62.55        & 62.91    & 62.70\\
\midrule
\multicolumn{5}{c}{\textbf{Prediction Horizon k = 100}} \\
\midrule
DeepLOB \citep{zhang2019deeplob}    &59.30       &58.12        & 59.30     & 57.88\\
DeepLOB-Seq2Seq     &59.74       &58.04        &59.74     & 58.45\\
DeepLOB-Attention   &60.00       & 59.56        & 60.00     &59.75\\
\bottomrule
\end{tabular}
\end{table}

\subsection{Comparison between IPU and GPU}

Anther important focus of this work is to utilise novel IPU hardward for the training process of our models. In particular, we want to benchmark training times using IPUs against equivalent times when using GPUs. In our experiment, we compare a single GPU (NVIDIA GeForce RTX 2080) to an IPU unit. We test, DeepLOB, DeepLOB-Seq2Seq, DeepLOB-Attention and other three networks separately on the GPU and IPU. The model training lasts for 200 epochs and we report the average training time per epoch in Table~\ref{table:train_time}.  

%\textcolor{red}{Also, 
%we compare the training time between a GPU and an IPU for DeepLOB-Attention with different input lengths in Figure~\ref{fig:train_time_input_len}.}

The IPU achieves superior performance in training speed and, in particular, it only takes about 15\% of the corresponding GPU wall-clock time to train an encoder-decoder model. 
The improvement in training speed is outstanding and it offers an alternative solution to deal with the slow training of an encoder-decoder network instead of using Transformers. 

\begin{table}[!t]
\centering
\caption{Average training time (per epoch) comparison between IPU and GPU.}
\label{table:train_time}
\begin{tabular}{l|l|l|l}
\toprule
\textbf{Model}             & \multicolumn{2}{l}{\textbf{Training Time} (in sec.)} & \textbf{\# of parameters} \\
\midrule
                  & GPU                    & IPU                   &                  \\
\midrule
MLP \citep{tsantekidis2017using}&  19            &   4 &  256515         \\
CNN-I \citep{tsantekidis2017forecasting} &   41                 &4                    &  17635          \\
LSTM \citep{tsantekidis2017using}  &  83                  &                   17 & 60099          \\
DeepLOB \citep{zhang2019deeplob}                                                     & 96                   &  15                  &  105347          \\
DeepLOB-Seq       & 215                   & 30                   & 176419           \\
DeepLOB-Attention & 270                   &33                   & 177699           \\
\bottomrule 
\end{tabular}
\end{table}

%\begin{figure}[!t]
%\centering
%\includegraphics[width=4.5in, height=2.3in]{train_time_input_len.pdf}
%\caption{Average training time (per epoch) for DeepLOB-Attention with different input lengths (lookback window size).}
%\label{fig:train_time_input_len}
%\end{figure}

%\textbf{\makecell{\% of\\+ Ret}}

%\begin{figure}[t]
%\centering
%\includegraphics[width=5.5in, height=2in]{correlation.pdf}
%\caption{Heatmap for rolling correlations between different index pair. (S: stock index, B: bond index, C: commodity index and V: volatility index.)}
%\label{fig:correlation}
%\end{figure}

\section{Conclusion}
\label{conclusion}

In this work we design multi-horizon forecasting models for limit order book (LOB) data by using deep learning techniques. We adapt encoder-decoder models,  Seq2Seq and Attention models, to generate forecast paths over multiple time steps. An encoder reads through the raw LOB data to extract representative features and a decoder steps through the output time step to generate multi-step forecasts. Our experiments suggest that our method delivers superior results compared to state-of-art algorithms. This is due to the iterative nature the decoder delivers which yields better predictive performance over long horizons as short-term estimates are fed into next prediction through an autoregressive structure. 

Encoder-decoder models rely on complex recurrent neural layers that often suffer from slow training processes. We address this problem by using a novel hardware IPU developed by Graphcore which is specifically designed for machine intelligence workload. We conduct a comparison between GPUs and IPUs to benchmark their training speed on modern deep neural networks for LOB data. We observe that IPUs leads to an acceleration that is significantly faster than common GPUs.  Such speed-ups in training time could open up a wide variety of applications, for example, application of online learning or reinforcement learning in the context of market-making, as such a high-frequency trading strategy has strict requirements on communication latency. It would be interesting to deploy IPUs to such setups and test their computational efficiency. Also, we can apply the encoder-decoder structure to a Reinforcement Learning framework as studied in \cite{zhang2020deep, zhang2020deepo}.

\section*{Acknowledgements}

The authors would like to thank Graphcore for making their hardware available for this study. They would also like to thank Alex Titterton and Alex Tsyplikhin from Graphcore for their valuable advise and help on hardware specific implementation details. Furthermore, the authors would like to thank the members of Machine Learning Research Group at the University of Oxford for their useful comments, as well as the anonymous referees for suggestions to improve the manuscript. Financial support from the Oxford-Man Institute of Quantitative Finance and Man Group is kindly acknowledged.

%\bibliographystyle{chicago}
%\bibliography{mybibliography.bib}

\appendix

\newpage
\section{Additional results on the LSE Dataset}
\label{train_lse}

\subsection{Training Configuration for the LSE Dataset}

\begin{figure}[H]
\centering
\includegraphics[width=4.5in, height=1.2in]{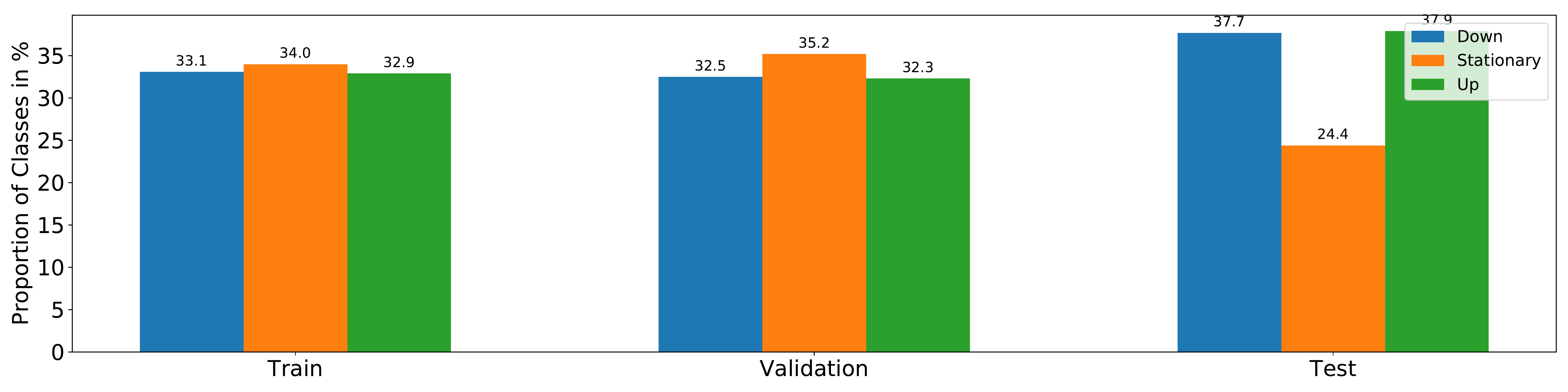}
\includegraphics[width=4.5in, height=1.2in]{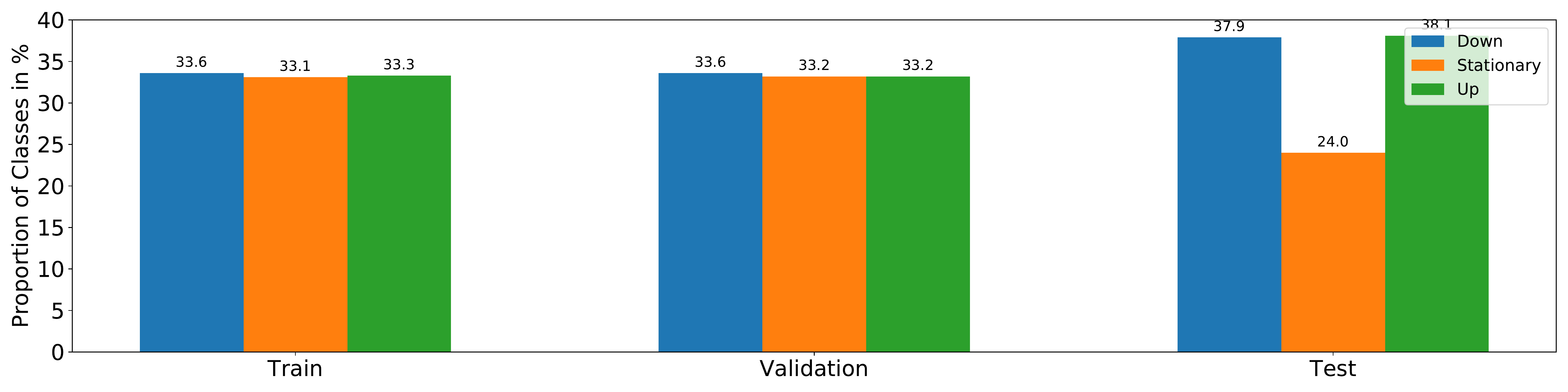}
\includegraphics[width=4.5in, height=1.2in]{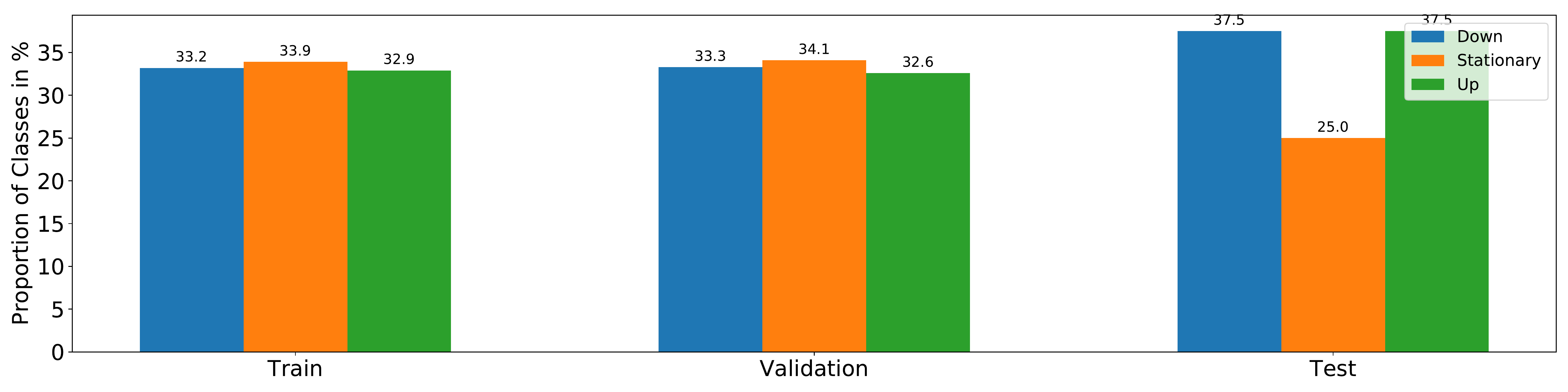}
\caption{Label class balancing for train, validation and test sets for different prediction horizons ($k$) for the LSE dataset. \textbf{Top:} $k=20$; \textbf{Middle:} $k=50$; \textbf{Bottom:} $k=100$.}
\label{fig:label_class}
\end{figure}

Following the setup in \cite{zhang2019deeplob, zhang2021deep},  we predict the future price movements into three classes: the market going up, staying stationary or going down.  Mid-prices are used to create labels and we define 
\begin{equation} \label{eq:label}
\begin{split}
&l_t = \frac{m_+(t) - m_-(t)}{m_-(t)}, \\
&m_-(t) = \frac{1}{k} \sum_{i=0}^{k-1} p_{t-i}, \\
&m_+(t) = \frac{1}{k} \sum_{i=1}^k p_{t+i},
\end{split}
\end{equation}
where $k$ is the prediction horizon and $p_t$ is the mid-price at time $t$.  We compare $l_t$ with a threshold ($\alpha$) to decide on the label,  and if $l_t>\alpha$,  we label it as up or $l_t<-\alpha$ is a down. We label everything else as the stationary class and the choices of $k$ and $\alpha$ are listed in Table~\ref{tb:setting} for all instruments in consideration. This choice results in roughly balanced classes as shown in Figure \ref{fig:label_class}.

\begin{table}[!htb]
\centering
\caption{Label parameters ($\alpha$) for different prediction horizons and instruments for the LSE dataset (units in $10^{-4}$).}
\begin{tabular}{l|llllll}
\toprule
        & LLOY & BARC & TSCO & BT   & VOD  \\
\midrule
k = 20  & 0.25 & 0.35 & 0.10 & 0.40 & 0.22 \\
k = 50  & 0.50 & 0.65 & 0.70 & 0.70 & 0.45 \\
k = 100 & 0.75 & 0.95 & 1.20 & 1.00 & 0.70 \\
\midrule
        & AAL & ANTO & AZN & BATS   &BDEV  \\
\midrule
k = 20  & 0.52 & 1.00& 0.30 & 0.40 & 0.85 \\
k = 50  & 0.97 & 1.85 & 0.62 & 0.80 & 1.70 \\
k = 100& 1.40 & 2.55 & 0.92 & 1.10 & 2.30 \\
\midrule
        & BKGH & BNZL & CCH & CNA   &DCC  \\
\midrule
k = 20  & 0.85 & 0.25 & 0.55 & 0.50 & 0.50 \\
k = 50  & 1.60 & 1.00 & 1.40 & 1.10 & 1.70 \\
k = 100& 2.30 & 1.60 & 1.95 & 1.70 & 2.70 \\
\midrule
        & EVRE & EXPN & FRES & IHG   &LAND  \\
\midrule
k = 20  & 1.20 & 0.50 & 1.10 & 0.55 & 0.65\\
k = 50  & 2.30 & 1.00 & 2.10 & 1.10 & 1.20 \\
k = 100& 3.00 & 1.55 & 3.05 & 1.60 & 1.65 \\
\midrule
        & NXT & LGEN & ULVR & RBS   &WPP  \\
\midrule
k = 20  & 0.70 &0.20 & 0.15 & 0.50 & 0.60 \\
k = 50  & 1.40 & 0.85 & 0.40 & 1.20 & 1.20 \\
k = 100& 2.10 & 1.30 & 0.60 & 1.70 & 1.70 \\
\bottomrule
\end{tabular}
\label{tb:setting}
\end{table}

\subsection{Detailed per instrument performance for transfer learning}

Tables \ref{table:lse_seq_tranfer} and \ref{table:lse_att_tranfer} show additional performance metrics for the experimental results for transfer learning on a per instrument basis.

\begin{table}[!t]
\caption{Experiment results of DeepLOB-Seq2Seq for the transfer learning on the LSE dataset.}
\label{table:lse_seq_tranfer}
\centering
\begin{tabular}{l|llllllllll}
\toprule
\multicolumn{11}{c}{\textbf{Prediction Horizon k = 20}}                                   \\
\midrule
             & AAL  & ANTO & AZN  & BATS & BDEV & BKGH & BNZL & CCH  & CNA & DCC \\
\midrule             
Acc.  &60.13      &61.70      &68.20      &65.11      & 64.05     & 64.00     & 67.59     & 67.91     &67.57     & 68.54    \\
Pre. &58.61      &60.32      &68.60      &64.61      &63.01      &62.99      &72.81      & 69.81     &69.26     &72.97     \\
Re.    &60.13      &61.70      &68.20      &65.11      & 64.05     &64.00      & 67.59     & 67.91     &67.57     & 68.54    \\
F1.       & 56.91     &58.88      &68.36      &64.77     &63.00      &62.93      & 68.02     & 68.39     & 68.00    & 68.68   \\
\midrule
             & EVRE & EXPN & FRES & IHG  & LAND & NXT  & LGEN & ULVR & RBS & WPP \\
             \midrule
Acc.  & 60.78     &66.45      &59.09      & 64.02     & 63.60     &66.29      & 67.88     &68.75      & 68.67    & 66.19    \\
Pre. & 58.84     &66.88      &57.32      &63.45      & 62.60     &66.07      & 73.21     &  70.99    & 71.48    & 66.03    \\
Re.    &60.78      &66.45      &59.09      &64.02      &63.60      &66.29      & 67.88     &68.75      & 68.67    & 66.19    \\
F1.       & 57.83     &66.63      &55.48      &63.66      & 62.81     &66.16      & 68.02     &  69.23    &  69.06   & 66.09   \\
\midrule
\multicolumn{11}{c}{\textbf{Prediction Horizon k = 50}}                                   \\
\midrule
             & AAL  & ANTO & AZN  & BATS & BDEV & BKGH & BNZL & CCH  & CNA & DCC \\
\midrule             
Acc.  & 60.19     &61.37      &64.49      &62.88      & 62.93     &62.96      & 64.42     & 64.65     &64.67     & 62.66    \\
Pre. & 57.85     &59.31      &63.61      &61.38      & 61.08     &61.15      & 65.81     & 64.40     & 64.70    &  64.28   \\
Re.    &60.19      &61.37      &64.49      &62.88      &62.93      &62.96      & 64.42     & 64.65     &64.67     & 62.66    \\
F1.       &56.58     & 58.35     &63.88      & 61.53     & 60.80     &61.25      &64.89      &  64.52    &64.68     &63.22    \\
\midrule
             & EVRE & EXPN & FRES & IHG  & LAND & NXT  & LGEN & ULVR & RBS & WPP \\
             \midrule
Acc.  &59.75      &64.37      &59.68      & 62.28     & 62.30     &64.25      & 62.70     &64.46      & 63.97    & 62.95    \\
Pre. &57.03      &63.67      &56.66      & 60.74     & 60.43     & 63.12     & 64.60     & 63.90     & 64.51    & 61.71    \\
Re.    &59.75      &64.37      &59.68      &62.28      &62.30      &64.25      & 62.70     & 64.46     & 63.97    & 62.95    \\
F1.       &55.92      &63.96      &55.74      & 61.07     & 60.76     & 63.46     & 63.30     & 64.09     & 64.20    & 61.97   \\
\midrule
\multicolumn{11}{c}{\textbf{Prediction Horizon k = 100}}                                   \\
\midrule
             & AAL  & ANTO & AZN  & BATS & BDEV & BKGH & BNZL & CCH  & CNA & DCC \\
\midrule             
Acc.  & 58.64     & 59.47     &60.17      &60.27      &60.53      &59.90      & 60.04     & 60.78     &60.91     & 58.30    \\
Pre. &  56.53    &57.54      & 58.67     &58.68      &58.38      &58.15      &59.08      & 59.80     &59.68     &57.15     \\
Re.    &58.64      &59.47      &60.17      & 60.27     &60.53      &59.90      & 60.04     & 60.78     &60.91     & 58.30    \\
F1.       &  56.46    & 57.77     &59.01      &59.11      &58.83      &58.53      &  59.40    & 60.18     &60.06     &57.51    \\
\midrule
             & EVRE & EXPN & FRES & IHG  & LAND & NXT  & LGEN & ULVR & RBS & WPP \\
             \midrule
Acc.  &58.46      &60.65      &58.06      & 59.18     &59.72      & 60.64     & 58.88     & 59.77     & 60.47    &59.84     \\
Pre. &56.04      &59.29      &55.70      &57.61      &58.17      &59.04      & 58.14     &  58.07    & 59.30    & 58.12    \\
Re.    &58.46      &60.65      &58.06      &59.18      &59.72      &60.64      & 58.88     & 59.77     & 60.47    &59.84     \\
F1.       & 56.33     &59.72      &55.91      &58.04      & 58.64     &59.47      & 58.44      & 58.42     &59.72     & 58.57   \\
\bottomrule
\end{tabular}
\end{table}

\begin{table}[!t]
\caption{Experiment results of DeepLOB-Attention for the transfer learning on the LSE dataset.}
\label{table:lse_att_tranfer}
\centering
\begin{tabular}{l|llllllllll}
\toprule
\multicolumn{11}{c}{\textbf{Prediction Horizon k = 20}}                                   \\
\midrule
             & AAL  & ANTO & AZN  & BATS & BDEV & BKGH & BNZL & CCH  & CNA & DCC \\
\midrule             
Acc.   & 61.14     & 62.57     & 68.09     & 65.39     &64.80      & 64.62     & 66.65     & 67.58     &67.26     &67.48     \\
Pre.  & 59.37     &  61.14    &68.31      &64.78      &63.67      &63.57      & 71.96     &69.52      & 68.85    & 71.84    \\
Re.    & 61.14     & 62.57     & 68.09     & 65.39     &64.80      &64.62          & 66.65     &  67.58    &67.26     &  67.48   \\
F1.  &  57.74    &  59.84    &68.18      & 64.96     &63.72      &63.50      & 67.19     &68.10      &67.71     & 67.76    \\
\midrule
             & EVRE & EXPN & FRES & IHG  & LAND & NXT  & LGEN & ULVR & RBS & WPP \\
             \midrule
Acc.  &61.81      & 66.96     &60.26      &64.63      &64.21      & 66.99     & 66.54     &67.96      &67.93     & 66.31    \\
Pre. &59.75      &67.36      &58.04      &63.96      &63.16      &66.75      & 72.25     &69.61      & 70.75    & 66.00    \\
Re.    & 61.81     &66.96      &60.26      &64.63      &64.21      & 66.99     &66.54      &67.96      &67.93     & 66.31    \\
F1.       &58.77      & 67.13     &56.59      & 64.19     &63.35      &66.85      &66.81      & 68.42     &68.38     & 66.11   \\
\midrule
\multicolumn{11}{c}{\textbf{Prediction Horizon k = 50}}                                   \\
\midrule
             & AAL  & ANTO & AZN  & BATS & BDEV & BKGH & BNZL & CCH  & CNA & DCC \\
\midrule             
Acc.   & 61.03     & 61.94     &64.40      &62.93      & 63.51     &63.40      & 63.13     & 64.42     &64.10     & 60.85    \\
Pre.  &  58.70    & 60.09     &64.64      &62.02      & 62.09     &62.24      &68.04      & 66.14     &66.16     &  66.80   \\
Re.    & 61.03     &61.94      & 64.40     &62.93      &63.51      & 63.40     &63.13      & 64.42     & 64.10    & 60.85    \\
F1.  &  58.13    &  59.81    &  64.50    & 62.30     &62.32      &62.55      & 63.95     &64.99      & 64.76    & 61.83    \\
\midrule
             & EVRE & EXPN & FRES & IHG  & LAND & NXT  & LGEN & ULVR & RBS & WPP \\
             \midrule
Acc.  &60.63      &64.19      &60.88      &62.43      &62.59      & 64.40     &  60.54    &64.22      & 63.09    & 63.00    \\
Pre. & 58.16     &65.21      &58.02      &61.64    &61.50      &64.48      &      67.41& 64.52     & 66.80    & 62.65    \\
Re.    &60.63      &64.19      &60.88      &62.43      &62.59      &64.40      &60.54      &64.22     &63.09     & 63.00    \\
F1.       &57.88      &64.60      &58.17      &61.92      & 61.86     &64.43      & 61.41     & 64.35     &63.95     &62.79    \\
\midrule
\multicolumn{11}{c}{\textbf{Prediction Horizon k = 100}}                                   \\
\midrule
             & AAL  & ANTO & AZN  & BATS & BDEV & BKGH & BNZL & CCH  & CNA & DCC \\
\midrule             
Acc.   & 58.80     & 59.72     &60.80      &60.43      &61.32     &60.70      & 59.89     &  60.90    &61.32     & 57.22    \\
Pre.  &  56.39    & 57.73     & 61.04     &59.54      &59.76      &59.49      & 64.07     & 63.47     &62.92     & 62.24    \\
Re.    & 58.80     & 59.72     &60.80      &60.43      &61.32     &60.70           &59.89      & 60.90     &61.32     & 57.22     \\
F1.  &  55.91    & 57.80     & 60.91     &59.88      &60.21      &59.84      & 60.71     &61.74      & 61.88    &58.10     \\
\midrule
             & EVRE & EXPN & FRES & IHG  & LAND & NXT  & LGEN & ULVR & RBS & WPP \\
             \midrule
Acc.  &58.79      & 60.87     &58.82      & 59.67     &60.00      &61.40      & 56.81     &60.58      & 59.32    & 60.38    \\
Pre. & 56.13     &61.61      &55.95        &58.66      &58.96      &61.22      &63.37      &60.47      & 63.37    &60.01     \\
Re.    &58.79      &60.87      &58.82      &59.67      &60.00      &61.40      &56.81      &60.58      & 59.32     &60.38     \\
F1.       & 56.40     &61.18      &56.35      &59.03      &59.35      &61.30      & 57.74     &60.52      & 60.36   &60.18    \\
\bottomrule
\end{tabular}
\end{table}

\end{document}